%% file: icme2023template.tex
\documentclass{article}
\usepackage{spconf,amsmath,epsfig}
\pdfoutput=1
\let\OLDthebibliography\thebibliography
\renewcommand\thebibliography[1]{
  \OLDthebibliography{#1}
  \setlength{\parskip}{0pt}
  \setlength{\itemsep}{0pt plus 0.3ex}
}
\usepackage{graphicx}
\usepackage{amsmath}
\usepackage{amssymb}
\usepackage{booktabs}
\usepackage{caption}
\usepackage{float} 
\usepackage{subcaption}
\usepackage{multirow}
\usepackage{wrapfig}
\usepackage[colorlinks,linkcolor=blue]{hyperref}
\begin{document}\sloppy

\def\x{{\mathbf x}}
\def\L{{\cal L}}

\newcommand\blfootnote[1]{%
\begingroup
\renewcommand\thefootnote{}\footnote{#1}%
\addtocounter{footnote}{-1}%
\endgroup
}

\title{Robust Cross-Modal Knowledge Distillation for Unconstrained Videos}
%

\name{
\textbf{Wenke Xia}\textsuperscript{1,2}, 
\textbf{Xingjian Li}\textsuperscript{2,3}, 
\textbf{Andong Deng}\textsuperscript{2,4}, 
\textbf{Haoyi Xiong}\textsuperscript{2}, 
\textbf{Dejing Dou}\textsuperscript{5},
\textbf{Di Hu}\textsuperscript{1,$\ast$}
}

\address{
\textsuperscript{1}Gaoling School of Artificial Intelligence, Renmin University of China\\
\textsuperscript{2}Big Data Lab, Baidu Research 
\textsuperscript{3}State Key Lab of IOTSC, University of Macau\\
\textsuperscript{4}CRCV, University of Central Florida \textsuperscript{5}BCG X
}
\maketitle

\blfootnote{$\ast$ Corresponding author.}



%
\begin{abstract}
\input{sections/abstract.tex}

\end{abstract}
\input{sections/Introduction.tex}

\input{sections/related.tex}
\input{sections/method.tex}

\input{sections/experiments.tex}

\vspace{-1em}
\section{Conclusion}
In this work, we point out that the irrelevant modality noise and differentiated semantic correlation between modalities, could affect the quality of cross-modal distillation performance for the multi-modal unconstrained data. To alleviate this issue, we propose a robust cross-modal distillation approach, with the modality noise filter module to explicitly filter irrelevant noise in teacher modality, and the contrastive semantic calibration module to distill useful knowledge for target task, leveraging the differentiated sample-wise semantic correlation. 
Extensive experiments show that our method could outperform other distillation methods on action recognition, video retrieval, as well as audio tagging tasks. 
\vspace{-1em}
\section{acknowledgement}
This research was supported by National Natural Science
Foundation of China (NO.62106272), the Young Elite Scientists Sponsorship Program by CAST (2021QNRC001), and
Public Computing Cloud, Renmin University of China.
\vspace{-1em}

\bibliographystyle{IEEEbib}
\bibliography{icme2023template}

\end{document}

%% file: sections/abstract.tex
Cross-modal distillation has been widely used to transfer knowledge across different modalities, enriching the representation of the target unimodal one. Recent studies highly relate the temporal synchronization between vision and sound to the semantic consistency for cross-modal distillation. However, such semantic consistency from the synchronization is hard to guarantee in unconstrained videos, due to the irrelevant modality noise and differentiated semantic correlation. To this end, we first propose a \textit{Modality Noise Filter} (MNF) module to erase the irrelevant noise in teacher modality with cross-modal context. After this purification, we then design a \textit{Contrastive Semantic Calibration} (CSC) module to adaptively distill useful knowledge for target modality, by referring to the differentiated sample-wise semantic correlation in a contrastive fashion. Extensive experiments show that our method could bring a performance boost compared with other distillation methods in both visual action recognition and video retrieval task. We also extend to the audio tagging task to prove the generalization of our method. The source code is  available at \href{https://github.com/GeWu-Lab/cross-modal-distillation}{https://github.com/GeWu-Lab/cross-modal-distillation}.

%% file: sections/Introduction.tex
\vspace{-1em}
\section{Introduction}

\label{sec:intro}


Knowledge distillation has become an effective approach to ensemble different data sources to enrich the representation ability of target modality \cite{gupta2016cross}. Compared with visual-related modalities (e.g. RGB and optical flow), sound naturally contains vivid supervisory information from an independent auditory source, which deserves thorough exploration due to its synchronized nature with vision. 
Inspired by this, cross-modal knowledge distillation has been extensively investigated in audio-visual learning  \cite{aytar2016soundnet}, leveraging the synchronization between auditory and visual modalities. 
\begin{figure}[t]
    \centering
    \includegraphics[width=0.47\textwidth]{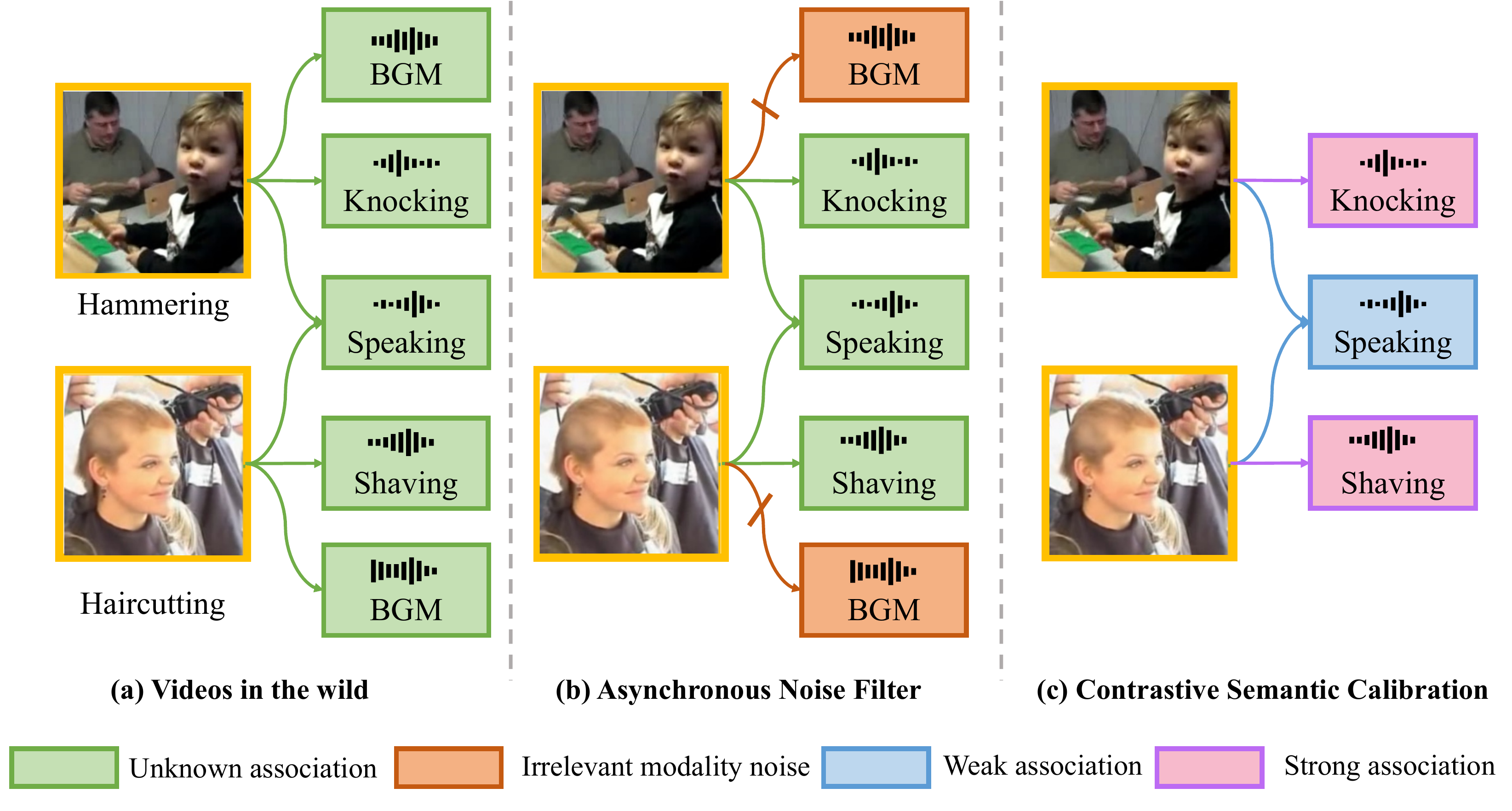}
    
    \caption{The illustration of our cross-modal distillation pipeline. (a) showcases unconstrained videos with irrelevant modality noise and differentiated semantic correlation. (b) shows that we first erase the irrelevant modality noise. (c) illustrates that we further capture the audio-visual correspondence and rectify the semantic correlation.}
    \label{fig:intro}
    \vspace{-2em}
\end{figure}

The aforementioned audio-visual distillation methods \cite{aytar2016soundnet} highly relate the cross-modal synchronization to semantic consistency, and directly transfer knowledge across the audio-visual modalities. However, such semantic consistency is hard to guarantee in unconstrained videos, thus, directly transferring knowledge across audio-visual modalities assuming highly-related semantic consistency could harm the distillation performance. The similar phenomenon of semantic misalignment is also found in  self-supervised multi-modal learning~\cite{morgado2021robust} but got little attention in cross-modal distillation.

In this work, we reconsider the conventional semantic consistency assumption and point out that the irrelevant modality noise and differentiated semantic correlation are blamed for the failure of such assumption and harm the cross-modal distillation performance, especially in unconstrained audio-visual scenarios.
Concretely, in cross-modal distillation, irrelevant modality noise is considered as the signals in one modality that are unrelated to its accompanying target modality. For instance, the BGM almost has no direct semantic correlation with the ``Hammering'' visual action in Figure~\ref{fig:intro}(b), thus transferring the knowledge in BGM would mislead the visual modality.
Meanwhile, considering the differentiated semantic correlation, the stronger the semantic correlation is, the more likely the visual event could be inferred from its accompanying sound. 
As demonstrated in Figure~\ref{fig:intro}(c), the speaking sound could widely relate to various visual human events, in contrast, the knocking sound almost co-occurs with the ``Hammering'' action. In this situation, the former case is regarded as weak semantic correlation, while the latter is referred to as strong semantic correlation.

Based on the consideration above, we propose to filter irrelevant modality noise first and then calibrate differentiated semantic correlation. 
Concretely, we first present a \textit{Modality Noise Filter} (MNF) module with a feature selection block to erase the irrelevant noise in teacher modality, leveraging a triplet loss to explicitly encourage the purified teacher to share more similarities with the student, which contributes to capturing related cross-modal knowledge.
Based on the purification, we further propose a \textit{Contrastive Semantic Calibration} (CSC) module to adaptively distill useful knowledge for target modality in a contrastive fashion, by referring to the differentiated sample-wise semantic correlation.
Considering that directly transferring cross-modal knowledge could deteriorate the quality of distillation~\cite{Ren_2021_CVPR}, we perform the above distillation only within student modality, but under the guidance of the teacher one.
We evaluate our method on different cross-modal distillation tasks and the experiments prove that our method could effectively transfer robust cross-modal knowledge for unconstrained videos.


To summarize, our contributions are as follows: (1) We focus on the cross-modal distillation for unconstrained videos and point out that the irrelevant modality noise and the differentiated semantic correlation, could seriously affect the quality of cross-modal knowledge distillation. (2) We propose a robust cross-modal distillation method with modality noise filter and contrastive semantic calibration modules, to erase irrelevant modality noise and adaptively distill useful knowledge for target modality leveraging the differentiated semantic correlation. (3) We provide qualitative and quantitative experiments, which show that our approach consistently outperforms other distillation methods in visual action recognition, video retrieval, as well as audio tagging tasks.

%% file: sections/related.tex
\vspace{-1em}
\section{Related Work}

\textbf{Cross-Modal Knowledge Distillation.}
To transfer knowledge across modalities, some earlier work \cite{gupta2016cross,aytar2016soundnet} applied distillation to transfer knowledge in the label space. Further, most frameworks \cite{tian2020contrastive} utilized feature-level distillation and obtained superior performance. Recent works \cite{Ren_2021_CVPR,chen2021distilling} pointed out that there exists semantic misalignment between modalities to different extents which harm the quality of cross-modal distillation, and they proposed fusing multi-modal features to reduce such semantic misalignment.
However, the correlation between modalities is practically complex beyond the simple case of misalignment, which needs more exploration.
In this work, we conclude the factors of such complex correlation come from the irrelevant modality noise and differentiated semantic correlation, based on which we propose to adaptively alleviate these issues for robust cross-modal distillation. \\
\textbf{Audio-Visual Learning.}
Recent progress in audio-visual learning has demonstrated its versatility in various application scenarios \cite{wei2022learning,xu2023mmcosine}, such as audio-visual question answering \cite{Li2022Learning} and sound source localization \cite{hu2021class}. Compared with visual-related modalities, sound provides useful information for perception, but also brings complicated problems when dealing with audio-visual modalities \cite{peng2022balanced}.
In this work, we focus on cross-modal distillation in unconstrained audio-visual scenarios, where the differentiated semantic correlation is addressed deliberately to ensure the robustness of distillation.



%% file: sections/method.tex
\begin{figure*}
    \centering
    \includegraphics[width=0.9\textwidth]{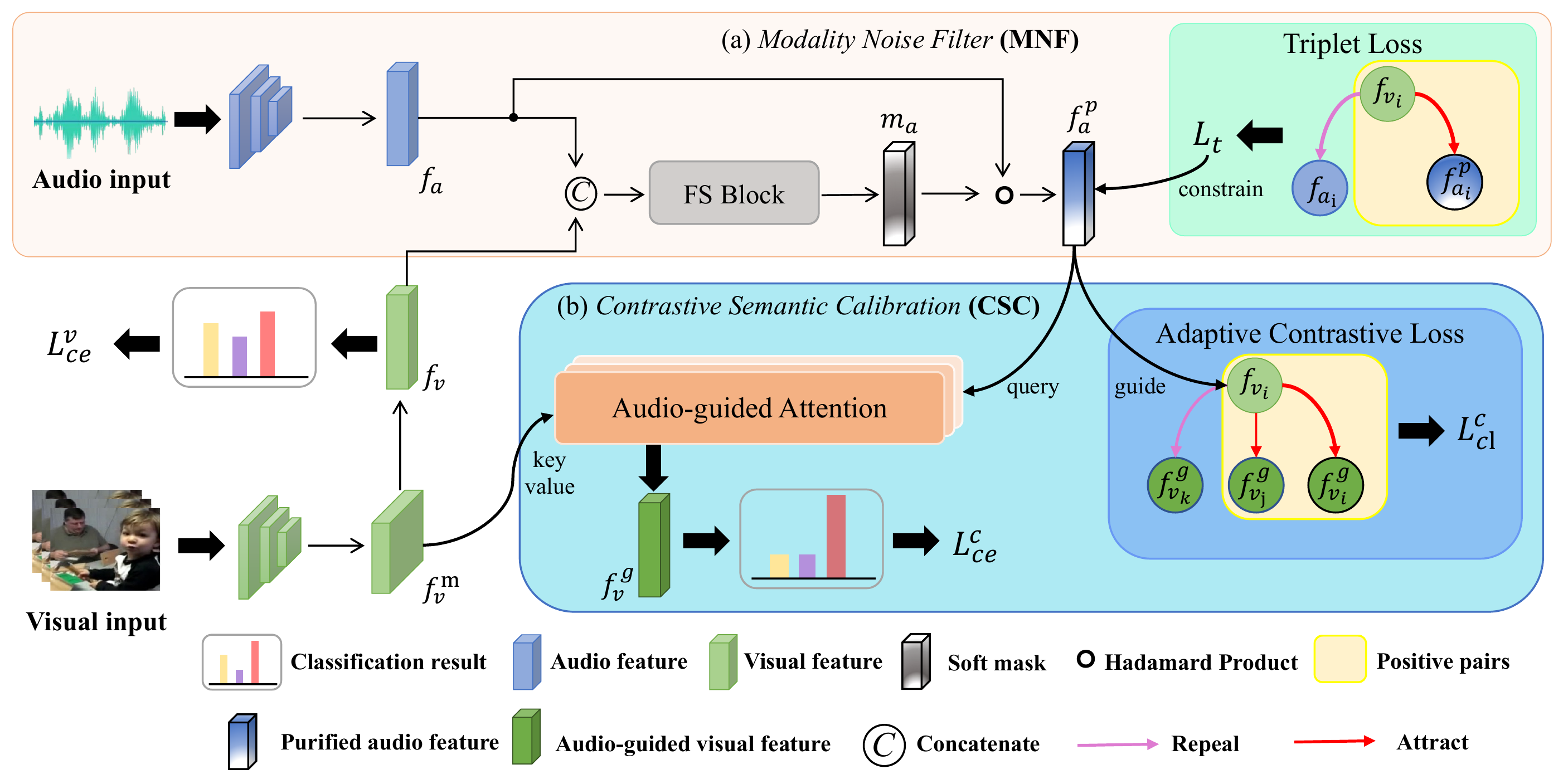}
   \caption{The pipeline of our robust cross-modal distillation approach. We first propose an \textit{Modality Noise Filter} module to erase the irrelevant noise in the auditory modality. Then, we present a \textit{Contrastive Semantic Calibration} module to adaptively distill useful knowledge for visual modality, referring to the sample-wise differentiated semantic correlation. In \textit{Contrastive Semantic Calibration} module, the thickness of the arrow indicates the semantic correlation strength between audio-visual pairs.}
    \label{fig:pipeline}
    \vspace{-1em}
\end{figure*}

\vspace{-1em}
\section{Method}

\input{sections/methods/Problem.tex}

\input{sections/methods/MNF.tex}

\input{sections/methods/CSC.tex}

\input{sections/methods/loss.tex}

%% file: sections/methods/Problem.tex
\subsection{Problem Formulation} \label{section:problem}

Given a dataset of $N$ samples as $ D = \{s_n,y_n\}_{n = 1}^N $, where $y_n$ is the category of the $n$-sample with $y_n \in \{1,...,K\} $, $K$ is the number of categories, and $s_n$ contains the audio recording and the visual clip $\{a_n,v_n\}$. 
Our goal is to enhance the representation of one modality with knowledge from the other one. Thus, the teacher and student modalities are both used for cross-modal distillation during training, while only the student modality is used for target tasks in inference time.
For simplicity, we take the auditory modality as teacher and the visual modality as student to present our approach\footnote{In experiments, we also show our method is flexibly employed to transfer knowledge from vision to sound.}.


\textbf{Audio Teacher Network.} Given the audio track $a_n$ of the video sample $s_n$, the spectrogram passes through a 2D-CNN pre-trained teacher model to obtain the audio feature $f_a$.

\textbf{Visual Student Network.}We borrow R(2+1)D \cite{tran2018closer} as the backbone to learn the visual representation from scratch. We input the visual clip $v_n$ and extract visual feature maps at the penultimate layer and take them as $f_{v}^m \in \mathbb{R}^{T' \times H' \times W' \times d_v}$, where $d_v$ is the number of the feature maps, and $T',H',W'$ are the temporal and spatial shape of each map. The visual feature $f_v$ is finally obtained through a global-average pooling on $f_{v}^m$, in both spatial and temporal dimensions.



%% file: sections/methods/MNF.tex
\vspace{-1em}
\subsection{Modality Noise Filter} \label{section:cross-attention}
As described above, there usually exists noise irrelevant in one modality to its accompanying modality in audio-visual scenarios.
Thus, we propose a \textit{Modality Noise Filter} (MNF) module, to explicitly erase the irrelevant noise in teacher modality with a feature selection block, encouraging the purified features in teacher modality to take higher semantic similarity with the student ones, as shown in Figure~\ref{fig:pipeline}(a).

Concretely, we present a channel-wise attention approach to filter irrelevant noise in teacher modality w.r.t. the content of student one.
Formally, we concatenate $f_v$ and $f_a$, then take them into a \textit{Feature Selection Block} (FS Block), which consists of nonlinear transformation and ends with a sigmoid function, to get soft mask $m_a$ and purified audio feature $f_{a}^p$:
\vspace{-0.5em}
\begin{equation} 
\label{audio_mask}
m_a = FS \left( concat[f_v,f_a]\right),
\end{equation}

\vspace{-2em}

\begin{equation}
    \label{visual_fuse}
    f_{a}^p = f_a \circ m_a ,
\end{equation}
where $\circ$ is a Hadamard product operation. We expect the soft mask $m_a$ could activate specific feature dimensions for different audio features, with the information of the visual modality. However, although such channel-wise attention has been proven effective in suppressing the importance of noisy features \cite{hu2018squeeze}, it fails to work in an implicit way as shown in our experiments, due to the serious difference between modalities in unconstrained videos. Thus, we propose the triplet loss $L_t$ to explicitly encourage the soft mask to erase irrelevant noise in the auditory modality, by drawing near the purified audio feature $f_a^p$ and the visual feature $f_v$:
\begin{equation}
    \label{triplet_loss}
    L_t = max(\phi(f_a,f_v) - \phi(f_a^p,f_v) + margin,0),
\end{equation}
where $\phi$ is the cosine similarity and $margin$ is a hyper-parameter to control the similarity degree. Especially, we detach the video feature $f_v$ when calculating the triplet loss $L_t$ so that only $m_a$ is learned to capture useful knowledge in teacher modality. Notably, we regard the original pre-trained audio feature $f_a$ as negative pair while the purified audio feature $f_a^p$ as positive pair in the triplet loss, to force the $f_{a}^p$ capture higher semantic similarity with the visual information $f_v$, which could help the soft mask $m_a$ to filter the irrelevant noise in teacher modality

%% file: sections/methods/CSC.tex
\vspace{-1em}
\subsection{Contrastive Semantic Calibration} \label{section:csc}

Even though the MNF module erases the irrelevant noise in teacher modality, the remained auditory and visual content still suffer from the differentiated semantic correlation, which would harm the distillation performance if we directly perform cross-modal distillation without proper adaption. Thus, we propose a \textit{Contrastive Semantic Calibration} (CSC) module to distill useful knowledge for student modality, referring to the differentiated semantic correlation in a contrastive fashion, as shown in Figure~\ref{fig:pipeline}(b).  

We introduce an audio-guided spatiotemporal attention mechanism $T_{attn}$, which takes purified audio feature $f_{a}^p$ as query and visual feature map $f_v^m$ as key and value, to capture the semantically consistent content in the visual modality. Additionally, a nonlinear transformation $T_{v}$ is deployed to get audio-guided visual feature $f_v^g$:
\begin{equation}
    \label{cross-attention}
    f_{v}^g = T_v\left( T_{attn}\left( f_{a}^p,f_{v}^m,f_{v}^m\right)\right).
\end{equation}
Considering that directly transferring cross-modal knowledge could deteriorate the quality of knowledge distillation \cite{Ren_2021_CVPR}, therefore, we propose to only transfer the audio-guided knowledge within visual modality, which has also been proven effective in our experiments. Thus, to encourage the audio-guided attention to focus on the useful knowledge within the visual modality, we optimize $f_v^g$ with cross-entropy loss $L_{ce}^c$ for the target task.

Contrastive learning has been widely used to distill knowledge in the feature space \cite{tian2019contrastive}. However, as demonstrated in Figure~\ref{fig:intro}, although the speaking and knocking sound both occur in the videos of ``Hammering'' action, they contain knowledge of inequable importance, which is ignored in traditional contrastive learning. Thus, we propose a sample-wise adaption method to calibrate the contrastive learning leveraging the differentiated semantic correlation:
\begin{small}
\begin{equation}
    L_{cl}^{c}= - \frac{1}{|B||P|}  \sum_{i \in B} \sum_{p \in P} log \frac{\exp (\phi(f_{v_i},f_{v_p}^g) / \tau_{ip} )}
    { \sum_{j\in B} \exp (\phi(f_{v_i},f_{v_j}^g)/ \tau_{ij})},
\end{equation}
\end{small}
where $B$ represents the set of samples in the same batch of $v_i$ and $P$ is the set of positive samples with the sample $f_i$. The temperature of each pair $\tau_{ij}$ is adaptively calculated as below:
\begin{equation}
    \label{eq:temp}
    \tau_{ij} = \phi(f_{v_i},f_{a_j}^p) + \delta,
\end{equation}
where $\delta$ is the hyper-parameter to control the degree of distillation. 
With such an adaption strategy, the stronger semantic correlation pairs would contain larger temperatures, which encourages these pairs to be closer and transfer more significant knowledge within student modality, under the guidance of audio-visual semantic correlation.

%% file: sections/methods/loss.tex
\vspace{-1em}
\subsection{Loss Functions} \label{section:loss}

We first propose the modality noise filter module, explicitly leveraging triplet loss $L_t$ to erase the irrelevant noise in teacher modality with the content of student one. 
Further, we present the contrastive semantic calibration module to extract useful knowledge benefit to the target task with supervisory cross-entropy loss $L_{ce}^c$, and 
then leverage the differentiated semantic correlation to adaptively rectify contrastive loss $L_{cl}^{c}$.
By combining these losses, we achieve robust cross-modal distillation for unconstrained videos:
\begin{equation}
    \label{distill}
    L_{distill} = L_t + L_{ce}^c + L_{cl}^{c}.
\end{equation}

It should be noted that visual feature $f_v$ is trained not only by the knowledge distillation, but also under a cross-entropy loss $L_{ce}^v$ for the target task:
\begin{equation}
    \label{matrix_adaption}
    Loss=  L_{ce}^v +  \lambda L_{distill} ,
\end{equation}
where $\lambda$ is a hyper-parameter to control the trade-off between the two objectives.

%% file: sections/experiments.tex
\vspace{-1em}
\section{Experiments}
\input{sections/experiments/setting.tex}
\vspace{-1em}
\subsection{Distilling Auditory Knowledge to Visual Modality}
\input{sections/experiments/results.tex}

\input{sections/experiments/ablation.tex}

\vspace{-1em}
\subsection{Distilling Visual Knowledge to Auditory Modality}

\input{sections/experiments/v2a.tex}

%% file: sections/experiments/setting.tex
\subsection{Experimental Settings}

\begin{table*}[h]
\centering
\setlength{\tabcolsep}{3mm}{
\begin{tabular}{c|cccc|cccc}
\toprule
\multirow{2}{*}{Method} & \multicolumn{4}{c|}{UCF51 } & \multicolumn{4}{c}{ActivityNet } \\ \cline{2-9} 
    & Acc  & R@1   & mAP@100 & mAP@500 
    & Acc    & R@1    & mAP@100 & mAP@500  \\ 
\midrule[0.7pt]
Uni-modal &   66.5  & 65.3  & 70.7 &  60.6 &    50.0    &     39.7   & 38.2 & 28.5 \\ 
KD  &   65.8   &   62.1   &  62.3  & 52.6  &    50.8    & 44.2   &   42.0 &  32.0   \\
CRD  &  67.2 &  65.0    &  70.4  & 62.5  &    50.8   &  42.9  &  39.6  & 28.9 \\
RKD  &   67.3   &   64.7     & 68.9  &  60.0  & 51.0    &  42.2     & 39.2 &  29.0   \\
xID  &   67.0   &  67.1      & 69.6  &  61.3  &    49.3& 41.8   & 43.5  &   31.4 \\
CCL  &   70.3   &   70.0   &  84.7 &  75.2   &    52.0    &  43.3  &  43.4   & 32.9  \\ 
KD-noise   &   68.6  &   64.4    & 66.4  & 57.2  &    51.7    &    45.1    & 43.1  & 32.8 \\
RC-MC  &   71.0   &   70.1    &  81.4 &  73.8  &   52.2 & 44.8  & 43.4  & 32.8    \\  
\midrule
Ours  &   \textbf{72.9}   & \textbf{72.3}   &  \textbf{91.4}   & \textbf{82.3}      &    \textbf{52.9}    &   \textbf{48.4}      &  \textbf{50.5} & \textbf{41.0} \\
\bottomrule
\end{tabular}}
\caption{Visual action recognition and video retrieval results on UCF51 \cite{soomro2012ucf101} and ActivityNet \cite{caba2015activitynet}.}
\label{tab:sota}
\vspace{-1em}
\end{table*}


We evaluate our method in two different scenarios: (1) \textbf{Distilling auditory knowledge to visual modality}. In this scenario, we borrow a R(2+1)D-18 \cite{tran2018closer} as the backbone of the student visual model, and a 2D-CNN14 \cite{kong2019panns} pre-trained on the AudioSet \cite{gemmeke2017audio} as teacher auditory model. We evaluate our method on visual action recognition and video retrieval tasks, in UCF51 \cite{soomro2012ucf101} and ActivityNet \cite{caba2015activitynet} datasets, which both cover a wide range of complex human daily activities. For the visual action recognition task, top-1 accuracy (\%) is measured on the video level. We also present the video retrieval results with R@K and mAP@K metrics in visual modalities to show the capacity of representation learning of our method. 
(2) \textbf{Distilling visual knowledge to auditory modality}. In this scenario,  both auditory student network and visual teacher network use the ResNet-50, and the visual model is pre-trained with ImageNet \cite{deng2009imagenet}. We evaluate our method on audio tagging task in Kinetics-Sound \cite{arandjelovic2017look} dataset. For audio tagging task, we utilize the top-1 accuracy (\%) on the video level.

%% file: sections/experiments/results.tex
\subsubsection{Experimental Results}
We first present the results of uni-modal training, where the backbone is trained with only visual inputs.
Then, we extend traditional uni-modal distillation methods (KD \cite{hinton2015distilling}, RKD \cite{park2019relational}, CRD \cite{tian2019contrastive}) to audio-visual cross-modal distillation.
We also compare our method with multimodal distillation approaches including xID \cite{morgado2021robust} and CCL \cite{chen2021distilling}. We further compare our method with the approach which distills knowledge from noisy data including KD-noise \cite{yuan2020revisiting} and RC-MC \cite{hu2021learning}.

As shown in Table~\ref{tab:sota}, our method could not only bring better action recognition performance but also obtain huge improvement for retrieval metrics, which proves the effectiveness of our cross-modal distillation method. Besides, among these distillation approaches, the methods which directly transfer cross-modal knowledge perform worse than other methods which either consider the modality semantic gap (CCL) or erase the cross-modal discrepancy (RC-MC). This phenomenon proves that considering the irrelevant noise and differentiated semantic correlation between modalities would benefit cross-modal distillation.

%% file: sections/experiments/ablation.tex
\vspace{-1em}
\subsubsection{Comparison to Other Distillation Strategies}
\begin{figure}
    \centering
    \includegraphics[width=0.47\textwidth]{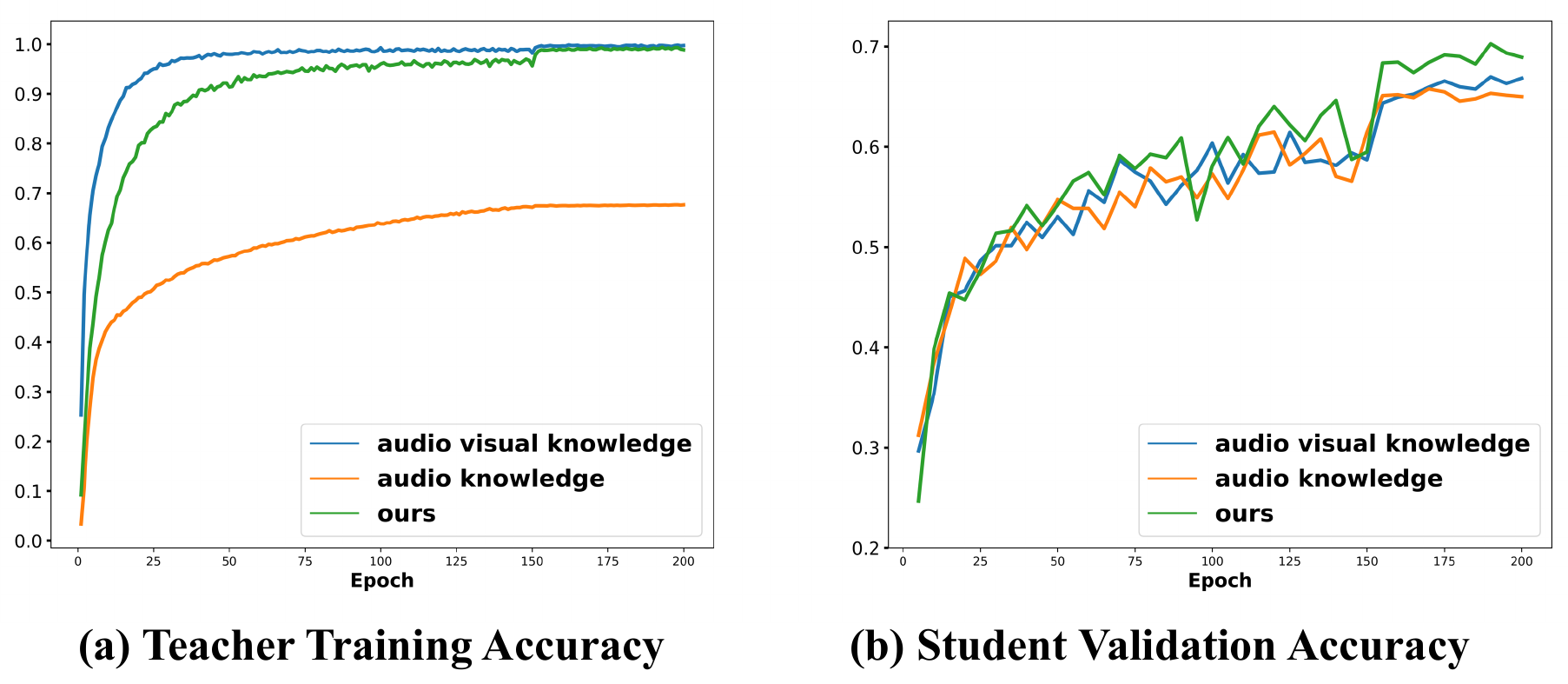}
    \caption{Comparative results for different distillation strategies. (a) demonstrates the training accuracy of the teacher model with multi-modal inputs, while (b) shows the uni-modal student performance in the validation set.}
    \label{fig:distill}
    \vspace{-1em}
\end{figure}

In this section, we provide qualitative experiments to prove the effectiveness of our distillation strategy that only transfers audio-guided knowledge within visual modality.
We introduce two different distillation strategies, where ``audio visual knowledge'' is the model which concatenates $f_a^p$ and $f_v^g$ to transfer multi-modal knowledge, while ``audio knowledge'' refers to the model which only transfers the original auditory modality knowledge $f_a$ to the visual student model.

As shown in Figure~\ref{fig:distill}, although the model with multi-modal knowledge could get higher accuracy during training, it fails to transfer more knowledge than our model with purified knowledge within visual modality, due to the multi-modal teacher contains lots of useless knowledge for student modality. Meanwhile, the teacher with only auditory knowledge fails to distinguish the visual events well even in the training set, which indicates that irrelevant noise in auditory modality seriously harms the model performance. However, our model could erase the irrelevant noise in teacher modality and capture useful knowledge within student modality, which gains significant improvement for the student model performance.

\vspace{-1em}
\subsubsection{Ablation Experiments}

\begin{table}[h]
\vspace{-1em}
\centering
\setlength{\tabcolsep}{2mm}{
\begin{tabular}{cc|cccc}
\multicolumn{2}{c|}{Methods}     & \multicolumn{4}{c}{Metrics} \\ 
\toprule

 $L_t$ & $L_{cl}^c $  & Acc   & R@1   & mAP@100 &mAP@500 \\ 
 \midrule

 $\times$ & $\times $ &  70.6 &  71.1   &  82.0  & 73.9 \\

 $\checkmark$ & $\times$ & 72.0 & 71.2  &  82.3  & 74.1 \\
 
 $\times$ & $\checkmark$ & 71.6 &  72.1  &  90.1  & 80.2 \\

 $\checkmark$ & $\checkmark$ & \textbf{72.9} &  \textbf{72.3}  &  \textbf{91.4}  & \textbf{82.3} \\

 \bottomrule
\end{tabular}}
\caption{Ablation results w.r.t the triplet loss $L_t$ and sample-wise adapted contrastive loss $L_{cl}^c$ on UCF51 dataset.}
\label{tab:anf}
\vspace{-1em}
\end{table}

We first provide the quantitative ablation results, where each method is combined with the audio-guided cross-entropy loss $L_{ce}^c$ to capture useful knowledge within student modality. As shown in Table~\ref{tab:anf}, the triplet loss $L_t$ and the sample-wise adaption contrastive loss $L_{cl}^c$ both bring improvement for the robust cross-modal distillation. 

\textbf{Modality Noise Filter.}
To validate the noise filter effect of our proposed MNF module, we mix the original audio recording with irrelevant noise, and transfer the knowledge from the mixed audio recording.
Concretely, for an input audio recording, we randomly mix with other audio recordings which belong to a different category. Then, we conduct experiments in two different settings, the first setting is that we remove the MNF module and directly pass the extracted feature to the audio-guided attention, and the other one is that we utilize the MNF module to filter irrelevant noise and pass the purified audio feature to the audio-guided attention. 

As shown in Table~\ref{tab:mix}, the results with our MNF module still remain comparable performance even with the audio mixed of noise (the accuracy only drops from 72.0 to 71.5 and the R@1 retrieval result even increases from 71.2 to 71.8, compared with the normal audio input as shown in the second line of Table~\ref{tab:anf}). However, the results without the MNF module decrease a lot in both the visual action recognition and video retrieval tasks. Besides, considering that only visual inputs are used for the student model during inference time, the mixed audio would not affect the student model, which could better validate the noise filter effect of our MNF module.

\begin{table}[t]

\centering
\setlength{\tabcolsep}{2mm}{
\begin{tabular}{c|cccc}
Methods     & \multicolumn{4}{c}{Metrics} \\ 
\toprule

   & Acc   & R@1   & mAP@100 &mAP@500 \\ 
 \midrule

 w/o MNF  &  69.3 &  69.2   &  78.2  & 70.2 \\

 w/ MNF &  \textbf{71.5} & \textbf{71.8}  &  \textbf{79.5}  & \textbf{72.3} \\

 \bottomrule
\end{tabular}}
\caption{Ablation results with mixed audio input.}
\label{tab:mix}
\vspace{-1em}
\end{table}

\begin{figure}
    \centering
    \includegraphics[width=0.47\textwidth]{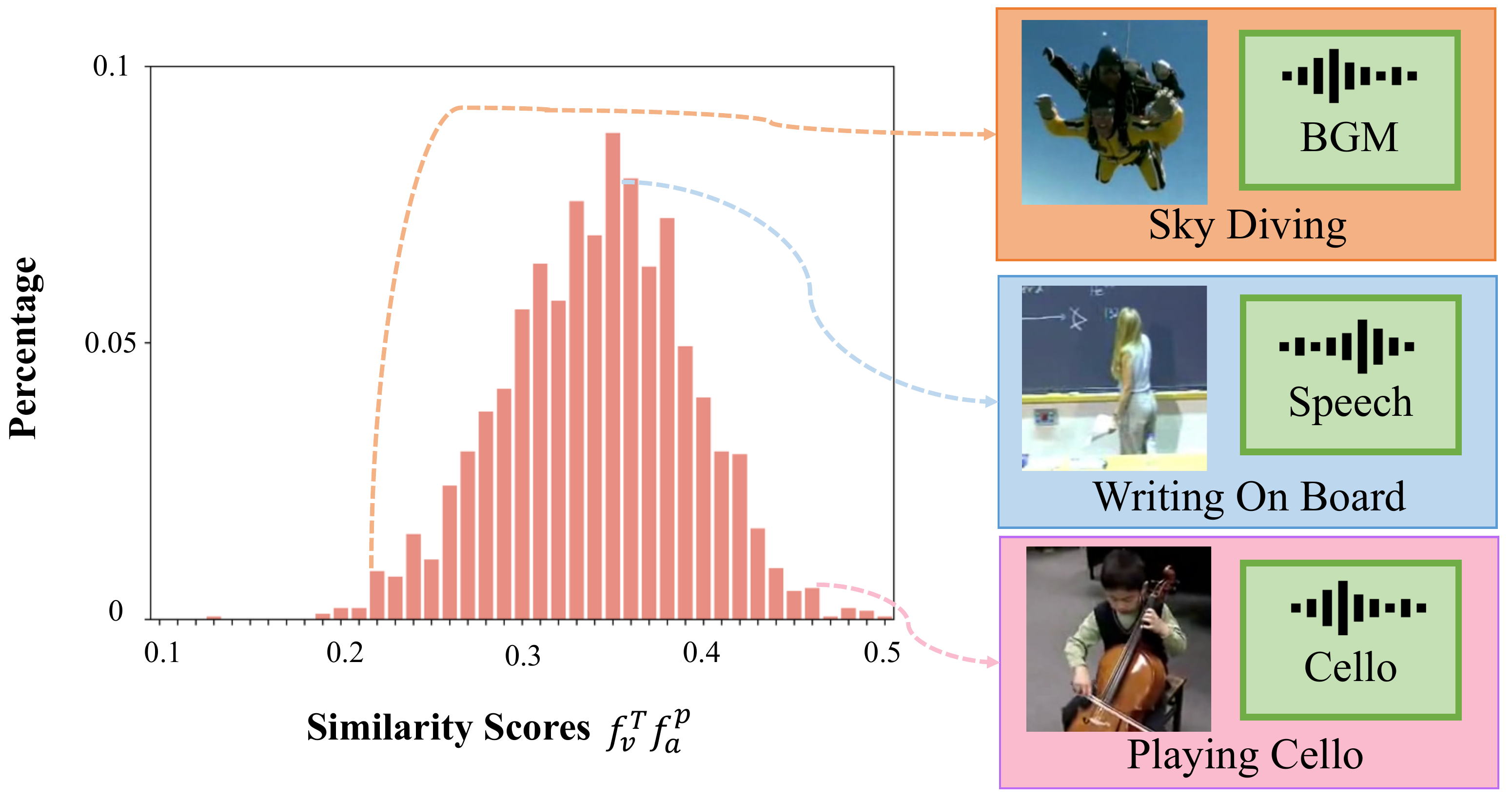}
    \caption{Audio-guided attention heatmap.}
    \vspace{-1em}
    \label{fig:sim}
\end{figure}

\textbf{Contrastive Semantic Calibration.}
As shown in Table~\ref{tab:anf}, our proposed CSC module could bring a huge improvement in video retrieval task. We point out it is the temperature adaption strategy with the sample-wise semantic correlation that brings robust fine-grained distillation performance and significant retrieval performance. As demonstrated in Figure~\ref{fig:sim}, our method could identify the semantic correlation reasonably to adaptively transfer useful knowledge. 

%% file: sections/experiments/v2a.tex
\begin{table}[]{}{}
\centering
\resizebox{0.45\textwidth}{!}{
\begin{tabular}{c|c|c|c|c|c|c}
\toprule
Method  & Uni-modal & KD & CRD & CCL & KD-noise & Ours \\ 
\midrule
Acc & 48.9 & 50.6 & 51.8 & 52.3 & 51.4 & \textbf{53.1} \\
\bottomrule

\end{tabular}
}
\caption{Audio tagging  results on Kinetics-Sound. }
\label{tab:v2a}
\vspace{-1em}
\end{table}

To demonstrate the generalization of our proposed method, we extend it to the audio tagging task. Concretely, we deploy a pre-trained image network as teacher model to distill the visual knowledge to auditory modality. ResNet is used as audio student network to follow the same attention mechanism as Figure~\ref{fig:pipeline} demonstrates.
Results shown in Table~\ref{tab:v2a} prove that our method outperforms the compared distillation methods, which presents the generalization of our method.